\def\year{2020}\relax
\documentclass[letterpaper]{article}
\usepackage{aaai20}
\usepackage{times}
\usepackage{helvet}
\usepackage{courier}
\usepackage[hyphens]{url}
\usepackage{graphicx}
\usepackage{graphicx}
\usepackage{amsmath,amssymb,amsfonts}
\usepackage{algorithmic}
\usepackage{textcomp}
\usepackage{xcolor}
\usepackage{subcaption}
\usepackage{booktabs}
\usepackage{siunitx}
\usepackage[acronym]{glossaries}
\usepackage{xspace}
\usepackage{multicol}

\title{
TOAD-GAN: Coherent Style Level Generation from a Single Example
}
\author{
Maren Awiszus, Frederik Schubert, Bodo Rosenhahn\\
Institut f\"ur Informationsverarbeitung\\
Leibniz University\\
Hanover, Germany \\
\{awiszus,schubert,rosenhahn\}@tnt.uni-hannover.de
}

\newcommand{\methodname}{TOAD-GAN\xspace}
\newcommand{\citet}[1]{\citeauthor{#1} \shortcite{#1}}
\newcommand{\citep}{\cite}

\newacronym{PCG}{PCG}{Procedural Content Generation}
\newacronym{GAN}{GAN}{Generative Adversarial Network}
\newacronym{PCGML}{PCGML}{PCG via Machine Learning}
\newacronym{EA}{EA}{Evolutionary Algorithm}
\newacronym{MCTS}{MCTS}{Monte Carlo Tree Search}
\newacronym{LSTM}{LSTM}{Long Short-Term Memory}
\newacronym{SAGAN}{SAGAN}{Self-Attention GAN}
\newacronym{WGAN-GP}{WGAN-GP}{Wasserstein GAN with Gradient Penalty}
\newacronym{DCGAN}{DCGAN}{Deep Convolutional GAN}
\newacronym{UMAP}{UMAP}{Universal Manifold Approximation and Projection}
\newacronym{TPKL-Div}{TPKL-Div}{Tile Pattern KL-Divergence}
\newacronym{ML}{ML}{Machine Learning}
\newacronym{FID}{FID}{Fr\'echet Inception Distance}
\newacronym{SMB}{SMB}{Super Mario Bros.}
\newacronym{RL}{RL}{Reinforcement Learning}

\begin{document}
\maketitle
\begin{abstract}

In this work, we present \textbf{\methodname} (\textbf{T}oken-based \textbf{O}ne-shot \textbf{A}rbitrary \textbf{D}imension \textbf{G}enerative \textbf{A}dversarial \textbf{N}etwork), a novel \gls*{PCG} algorithm that generates token-based video game levels.
\methodname follows the SinGAN architecture and can be trained using only one example.
We demonstrate its application for \emph{\acrlong*{SMB}} levels and are able to generate new levels of similar style in arbitrary sizes.
We achieve state-of-the-art results in modeling the patterns of the training level and provide a comparison with different baselines under several metrics.
Additionally, we present an extension of the method that allows the user to control the generation process of certain token structures to ensure a coherent global level layout.
We provide this tool to the community to spur further research by publishing our source code.

\end{abstract}
\glsresetall

\section{Introduction}
\label{sec:introduction}

Level design is a key component of the game creation process.
The designer has to consider aspects like physics, playability and difficulty during the creation process.
This task can require a lot of time for even a single level.

\gls*{PCG} has the potential to assist the designer by automating parts of the process.
Early works in \gls*{PCG} used the co-occurrence of tokens (e.g. a single enemy or ground block) in existing game levels to identify patterns \cite{dahlskogPatternsProceduralContent2012} and combined them using simple statistical models \cite{snodgrassGeneratingMapsUsing2014}.
The quality of these algorithms critically depends on the extracted patterns and co-occurrence relations, which have to be defined manually.
Recent approaches used \gls*{PCGML} \cite{summervilleProceduralContentGeneration2018} to learn the patterns and relations from the data automatically \cite{summervilleSuperMarioString2016,volzCapturingLocalGlobal2020}.
However, simply applying them to the level generation task comes with several drawbacks.
The Machine Learning algorithms need many examples to extract the patterns.
As manual level design is a costly process, there is usually a very limited number available for training.
Even if the amount is sufficient, the generated levels are mixtures of all example levels and do not have a coherent style.
Finally, most recent \gls*{PCGML} algorithms are black boxes and do not allow for high-level control of their generated content.

\begin{figure}[t!]
\begin{center}
\includegraphics[width=\linewidth]{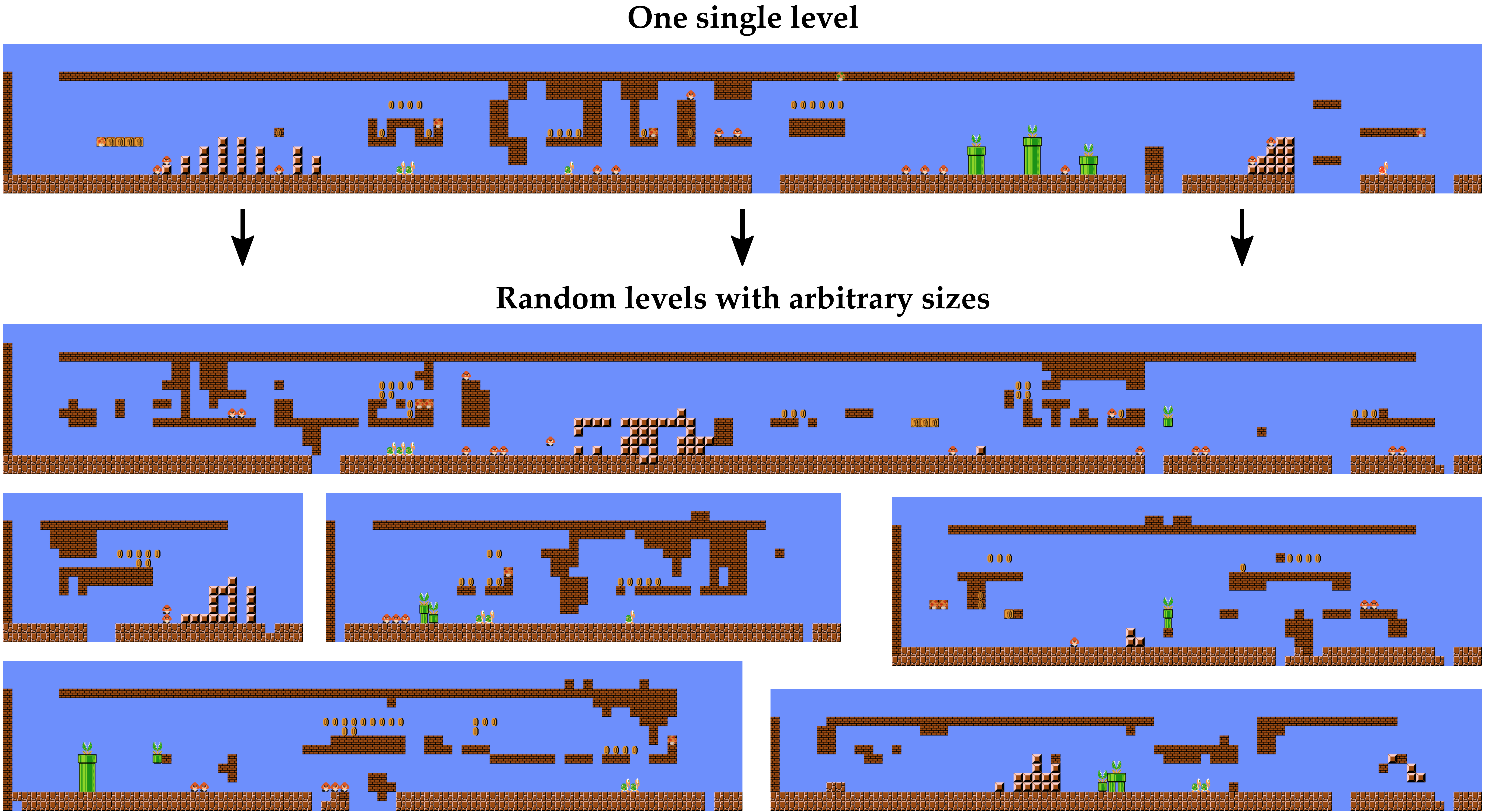}
\end{center}
   \caption{
   Generated levels based on \acrlong*{SMB} level 1-2.
   The new levels can be generated in any size and preserve the token distribution of the original level, while also containing new, previously unseen patterns.
   }
\label{fig:teaser}
\end{figure}

In this paper, we introduce \methodname as a solution to these problems.
Our work is inspired by SinGAN \cite{shahamSinGANLearningGenerative2019}, a recent \gls*{GAN} \cite{goodfellow2014generative} architecture that learns a generative model given only one example image.
This is achieved by learning patch-based features on different spatial scales.
Since SinGAN was developed for natural RGB images, it is unable to generate convincing video game levels that are based on 2D token maps.
Our method, on the other hand, was developed exactly with this purpose in mind.
As shown in Fig.~\ref{fig:teaser}, \methodname is able to generate new levels in the style of one given example level.

In summary, our \textbf{contributions} are:
\begin{itemize}
    \item With \methodname, we present a novel generative model that allows for level generation following the one-shot training approach of SinGAN.
    \item We introduce a downsampling algorithm for token-based levels specifically designed to preserve important tokens.
    \item An extension of our method enables authoring of the global level structure. We show this for example levels generated for Super Mario Kart.
    \item We visualise our generated content in a latent space to compare it with the original levels.
    \item We enable further research by publishing our source code.\footnote{https://github.com/Mawiszus/TOAD-GAN}
\end{itemize}

\section{Related Work}
\label{sec:related-work}

Generating \gls*{SMB} levels was one of the first challenges proposed to the \gls*{PCG} community \cite{shaker2010MarioAI2011}.
Since its introduction, many approaches were presented that tried to capture the patterns from the original levels and combine them in novel ways.
This section only covers a selection, due to the vast amount of \gls*{PCG} approaches.
For a review of pattern-based level generators for \gls*{SMB} see \citet{khalifaIntentionalComputationalLevel2019}.

\subsection{\acrlong*{SMB} Level Generation}
\label{subsec:mario_level_generation}

\citet{dahlskogPatternsProceduralContent2012} identified and analyzed patterns with different themes, such as enemies, gaps or stairs.
They assessed the difficulty of the patterns to human players and outlined how those patterns could be combined and varied to create new levels.
In their continued work \cite{dahlskogMultilevelLevelGenerator2014}, they additionally defined micro- (vertical slices) and macro-patterns (sequences of patterns).
Using an \gls*{EA}, they generated levels by selecting micro-patterns, with a fitness function based on the occurrence of patterns and macro-patterns.

Search-based \gls*{PCG} \cite{togeliusSearchBasedProceduralContent2011} was applied to \gls*{SMB} by \citet{summervilleMCMCTSPCGSMB2015}.
The authors used \gls*{MCTS} \cite{coulomEfficientSelectivityBackup2007} to guide the sampling process from a Markov Chain model of tokens.
The reward of the \gls*{MCTS} was computed based on the solvability, number of gaps, number of enemies and rewards (coins or power-ups).

\subsection{Neural Networks for PCG}
\label{subsec:networks-for-level-gen}

Recent \gls*{PCGML} approaches that use Neural Networks have also been applied extensively to the \gls*{SMB} level generation problem.
\citet{hooverComposingVideoGame2015} trained a neural network with an \gls*{EA} to generate new levels.
The network predicts the height of a token in a level slice, given the heights of all tokens in the previous slices.

\citet{summervilleSuperMarioString2016} trained their model on levels by predicting the tokens sequentially.
They used a neural network architecture based on \gls*{LSTM} cells \cite{hochreiterLSTM1997} to predict the next token, given a context of previous tokens in the unrolled level.

Recently, \glspl*{GAN} were used to create \gls*{SMB} levels.
In \cite{volzEvolvingMarioLevels2018b} the authors train a \gls*{GAN} on slices of the original levels and use an \gls*{EA} to search the space of generated levels by scoring the fraction of enemy and ground tokens.

A similar approach was taken by \citet{torradoBootstrappingConditionalGANs2019} who used the \gls*{SAGAN} \cite{zhangSelfAttentionGenerativeAdversarial2019} architecture and conditioned their generation process on a feature vector that contained the targeted token distributions of the generated levels.
This conditioning increased the variability of their generated content.

\begin{figure}
\begin{center}
\includegraphics[width=\linewidth]{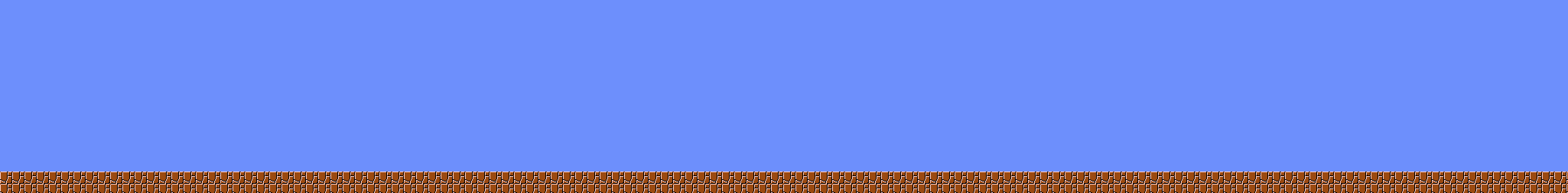}
\end{center}
   \caption{
   Example of the mode collapse of a \acrshort*{WGAN-GP} trained solely on $16 \times 16$ slices from \gls*{SMB} level 1-1. This slice size is chosen because square images are preferable for \glspl*{GAN} and 16 is the default height of all \gls*{SMB} levels.
   The generator stops creating anything but the ground and sky.}
\label{fig:mode_collapse}
\end{figure}

\section{Method}
\label{sec:method}

\begin{figure*}[ht!]
\begin{center}
\includegraphics[width=.97\linewidth]{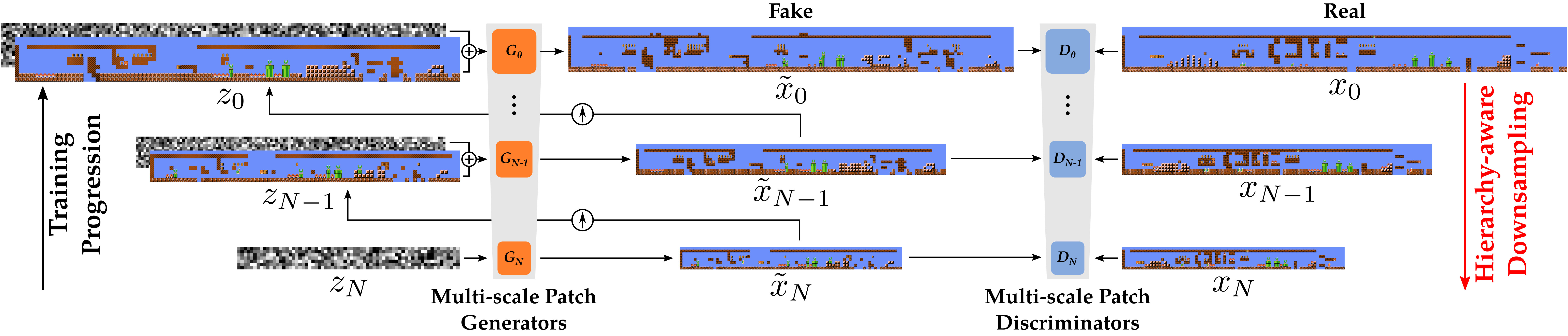}
\end{center}
   \caption{
   Generation process of \methodname on \acrlong*{SMB} level 1-2.
   The architecture is adapted from SinGAN (cf. Fig.~4 of \cite{shahamSinGANLearningGenerative2019}).
   We use a downsampling method on a one-hot encoded version of the level that preserves small but important structures which would be lost when performing simple spatial downsampling. The upwards arrow between the scales represents bilinear upsampling.
   }
\label{fig:base_pipeline}
\end{figure*}

Limited training data is one of the key problems of \gls*{PCGML} algorithms \cite{torradoBootstrappingConditionalGANs2019,bontragerFullyDifferentiableProcedural2020a}.
Therefore, the goal of our work is the generation of new levels for \gls*{SMB} from very little training data.
With \methodname, we take this problem to the extreme regime of learning from only one single training level.
Similar to other recent publications \cite{volzEvolvingMarioLevels2018b,torradoBootstrappingConditionalGANs2019,volzCapturingLocalGlobal2020}, \methodname is based on the \gls*{GAN} architecture.

\subsection{Generative Adversarial Networks}
\label{subsec:gans}

\glspl*{GAN} are able to generate samples from a given training distribution \cite{goodfellow2014generative}.
They consist of two adversaries: The generator $G$ maps random noise vectors $z$ to samples $\tilde{x}$, which the discriminator $D$ is trying to distinguish from real samples $x$.
In the end, $G$ produces $\tilde{x}$ that are indistinguishable from real $x$.
However, this process can become unstable.
In the low-data regime, the discriminator might be able to memorize the distribution of real samples and stops providing useful gradients for the generator.
Many different extensions to the basic architecture were proposed to stabilize the training process,
for example minimizing the Wasserstein distance \cite{arjovskyWassersteinGAN2017} and penalizing the norm of the gradients of the discriminator \cite{gulrajaniImprovedTrainingWasserstein2017a}.
The resulting \gls*{WGAN-GP} is able to model a variety of distributions, but it is still prone to failures like mode collapse.
This is the case when the generator produces samples that contain only a few features (or modes) of the data which the discriminator cannot classify correctly.
Then, the generator will never learn to produce the missed modes.
See Fig.~\ref{fig:mode_collapse} for an example.

Even though \glspl*{GAN} have shown promising results, their success depends on the availability of a lot of samples from the real distribution.
Additionally, long range correlations in an image can only be modeled by convolutional \glspl*{GAN} with many layers.
For longer levels, this increases the number of parameters that have to be optimized, which further complicates the training process.

\subsection{SinGAN}
\label{subsec:singan}

SinGAN \cite{shahamSinGANLearningGenerative2019} is a novel \gls*{GAN} architecture that enables learning a generative model from a single image.
This is achieved by using a cascade of generators and discriminators that act on patches from differently scaled versions of the image.
The weights of the models at each scale are initialized with those from the scale below.
This initialization bootstraps the training of the \glspl*{GAN} in upper scales and stabilizes the training process.

To generate a new sample, a noise map $z_N$ of arbitrary size is fed into the generator $G_N$ at the lowest scale.
At each subsequent scale $N < n \leq 0$, the output $\tilde{x}_{n+1}$ of the previous generator $G_{n+1}$ is up-sampled ($\uparrow$) and added to a new random noise map $z_n \sim \mathcal{N}(0, \sigma^2)$.
The variance $\sigma^2$ controls how much information from the lower scales is passed through to the upper layers of SinGAN.
Both are used by the generator $G_n$ to produce an output $\tilde{x}_n$ for the current scale
\begin{equation}
    \tilde{x}_n = \tilde{x}_{n+1}\uparrow + G_n (z_n + \tilde{x}_{n+1}\uparrow).
\end{equation}
The discriminators receive either a scaled real image or the output of their respective generator.
Generator and discriminator only act on patches and are fully-convolutional.
This means that the size of the output is determined by the size of the initial noise map at the lowest scale.


For a more in-depth explanation please refer to the original SinGAN paper by \citet{shahamSinGANLearningGenerative2019}.

\subsection{\methodname}
\label{subsec:singan_for_pcg}

Fig.~\ref{fig:base_pipeline} shows the pipeline of \methodname for the generation of \gls*{SMB} levels.
There are 15 original \gls*{SMB} levels provided by the Video Game Level Corpus \cite{summervilleVGLCVideoGame2016}, each with different characteristics.
The levels are placed in three worlds (overworld, underground, floating platforms) with different global structure and token patterns.
For training, one level is sampled down to $N$ different scales.
We choose $N$ such that the receptive field of the convolutional filters in our generators and discriminators is able to cover at minimum half of the height of the levels at the lowest scale.
This ensures that larger structures are modeled correctly, but allows for variation in their global position.

\begin{table}
    \centering
    \caption{Token Hierarchy}
    \begin{tabular}{l l l}
    \toprule
        ~ & Group & Tokens \\
    \midrule
        0 & Sky & \includegraphics[height=0.3cm]{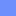} \\
        1 & Ground & \includegraphics[height=0.3cm]{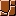} \\
        2 & Pyramid & \includegraphics[height=0.3cm]{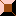}\\
        3 & Platforms &  
        \includegraphics[height=0.3cm]{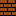}, 
        \includegraphics[height=0.3cm]{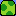}, 
        \includegraphics[height=0.3cm]{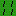},
        \includegraphics[height=0.3cm]{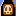}, 
        \includegraphics[height=0.3cm]{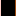} \\
        4 & Pipes & 
        \includegraphics[height=0.3cm]{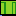}\includegraphics[height=0.3cm]{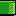}, 
        \includegraphics[height=0.3cm]{table01_09_pipe_top_left.png}\includegraphics[height=0.3cm]{table01_10_pipe_top_right.png} 
        + \includegraphics[height=0.3cm]{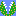} \\
        5 & Enemies &  
        \includegraphics[height=0.3cm]{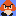}, 
        \includegraphics[height=0.3cm]{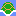}, 
        \includegraphics[height=0.3cm]{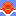}, 
        \includegraphics[height=0.3cm]{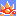}\\
        6 & Special Enemies & 
        \includegraphics[height=0.3cm]{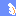}\includegraphics[height=0.3cm]{table01_13_gkoopa.png}, 
        \includegraphics[height=0.3cm]{table01_16_wings.png}\includegraphics[height=0.3cm]{table01_14_rkoopa.png}, 
        \includegraphics[height=0.3cm]{table01_16_wings.png}\includegraphics[height=0.3cm]{table01_15_spiky.png}, 
        \includegraphics[height=0.3cm]{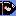}, 
        \includegraphics[height=0.3cm]{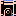} \\
        7 & Special Blocks &  
        \includegraphics[height=0.3cm]{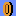}, 
        \includegraphics[height=0.3cm]{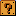} + \includegraphics[height=0.3cm]{table01_19_coin.png}, 
        \includegraphics[height=0.3cm]{table01_20_block_question.png} + \includegraphics[height=0.3cm]{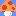}, 
        \includegraphics[height=0.3cm]{table01_04_block_brick.png} + \includegraphics[height=0.3cm]{table01_19_coin.png} \\
        8 & Hidden Blocks & 
        \includegraphics[height=0.3cm]{table01_04_block_brick.png} + \includegraphics[height=0.3cm]{table01_21_mushroom.png}, 
        \includegraphics[height=0.3cm]{table01_04_block_brick.png} + \includegraphics[height=0.3cm]{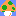}, 
        \includegraphics[height=0.3cm]{table01_01_sky.png} + \includegraphics[height=0.3cm]{table01_19_coin.png}, 
        \includegraphics[height=0.3cm]{table01_01_sky.png} + \includegraphics[height=0.3cm]{table01_22_1-Up.png} \\
        \hfill
    \end{tabular}
\label{tab:token_groups}
\end{table}

Interpreting each token as one pixel of an image and then downsampling naively results in lost information, as aliasing would make important tokens disappear at lower scales. 
To keep most of the information from the original level, we propose a downsampling method which preserves important tokens.
This method is inspired by TF-IDF weighting \cite{manning2008introduction} in Natural Language Processing where the importance of a term is defined by its term frequency multiplied by its inverse document frequency.
In our case, terms are tokens and documents are levels.
Tokens that occur often and in multiple levels, like the sky and ground blocks, are of lower importance than rare tokens, such as the hidden and special blocks.
The complete token hierarchy can be found in Tab.~\ref{tab:token_groups}.

The steps of this process are as follows.
First, bilinear downsampling is used on the one-hot encoded training level to create the base levels of the chosen scales.
For each pixel in each scale, the tokens with a value greater than zero are selected.
From that list, the tokens with the highest rank in our hierarchy are kept and the  remaining tokens are set to zero.
Finally, a Softmax is applied over all channels per pixel.
In Fig.~\ref{fig:base_pipeline} on the right, two downsampled versions of level 1-2 can be seen.
Later, we also need to sample the outputs on lower scales up.
For this we use bilinear up-sampling.

On natural images, SinGAN uses zero-centered gaussian spatial noise that is constant over all color channels for a given pixel, i.e. it only changes the brightness of that pixel.
This places a prior on the hue of the up-sampled pixels and increases the similarity of the generated samples between the different scales.
In our case, the channels represent the tile types.
Because these are independent from each other, we apply the noise to all channels individually.

\methodname can be extended to perform level authoring by injecting a predefined input into the generator cascade.
The generators fill in the details and produce a sample which follows the structure of the injected input but has a similar style as the training sample.
This application is particularly interesting for \gls*{PCG} as the designer can describe a desired level or layout for a given token and the generators create variants of it.
In our experiments, we inject a new, differently structured map for a specific token after the very first generation step.
This basic structure is preserved and expanded upon by the following generator steps, which results in a level with the desired structure that consists of the patterns learned by the generators.

\section{Experiments}
\label{sec:experiments}

\begin{figure}[t!]
\begin{center}
\textbf{\small{Overworld}}\\
\vspace{0.1cm}
\begin{subfigure}[b]{0.49\linewidth}
    \includegraphics[width=\linewidth]{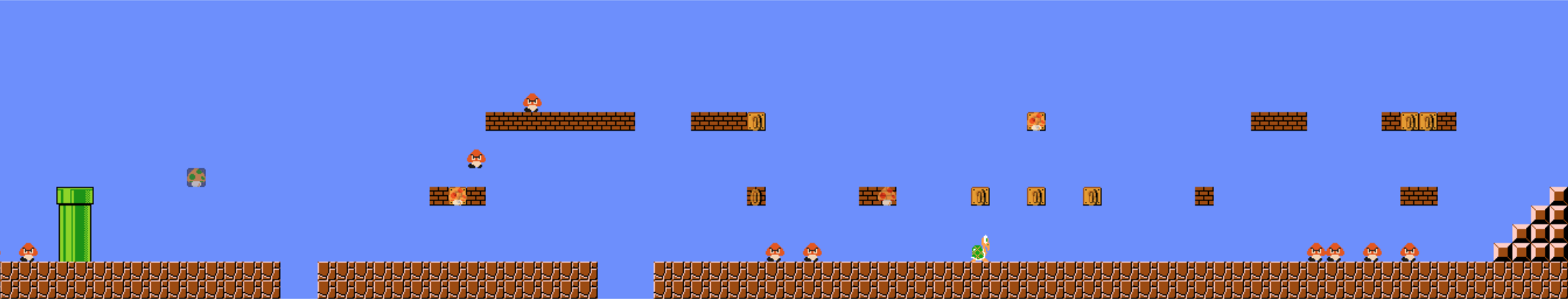}
    \caption{1-1 (original)}
\end{subfigure}
\hfill
\begin{subfigure}[b]{0.49\linewidth}
    \includegraphics[width=\linewidth]{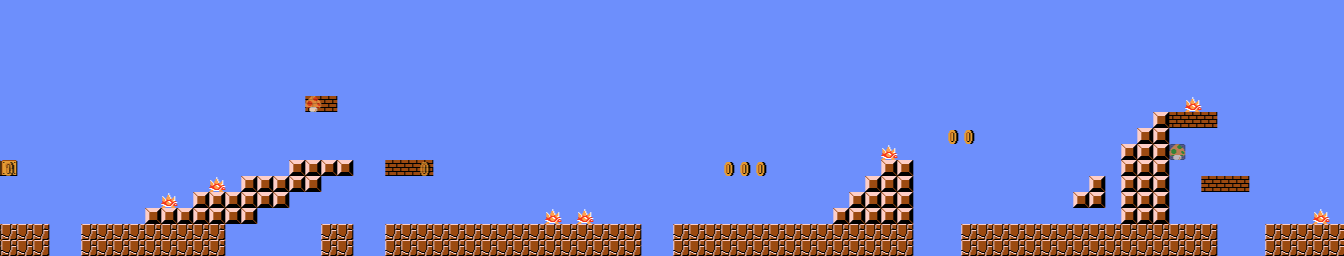}
    \caption{6-1 (original)}
\end{subfigure}\\
\begin{subfigure}[b]{0.49\linewidth}
    \includegraphics[width=\linewidth]{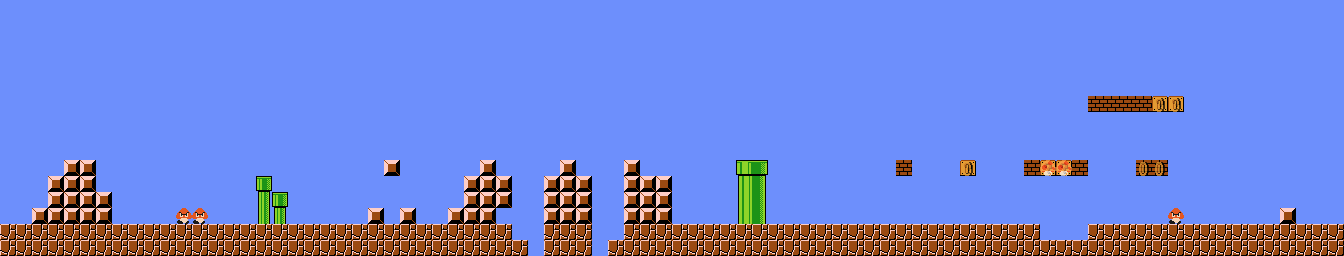}
    \caption{G 1-1 (ours)}
\end{subfigure}
\hfill
\begin{subfigure}[b]{0.49\linewidth}
    \includegraphics[width=\linewidth]{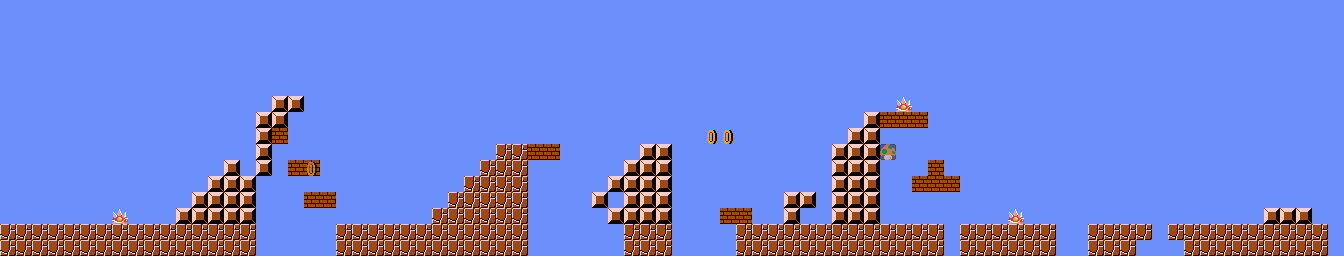}
    \caption{G 6-1 (ours)}
\end{subfigure}
\vspace{0.1cm}
\textbf{\small{Underground}}\\
\begin{subfigure}[b]{0.49\linewidth}
    \includegraphics[width=\linewidth]{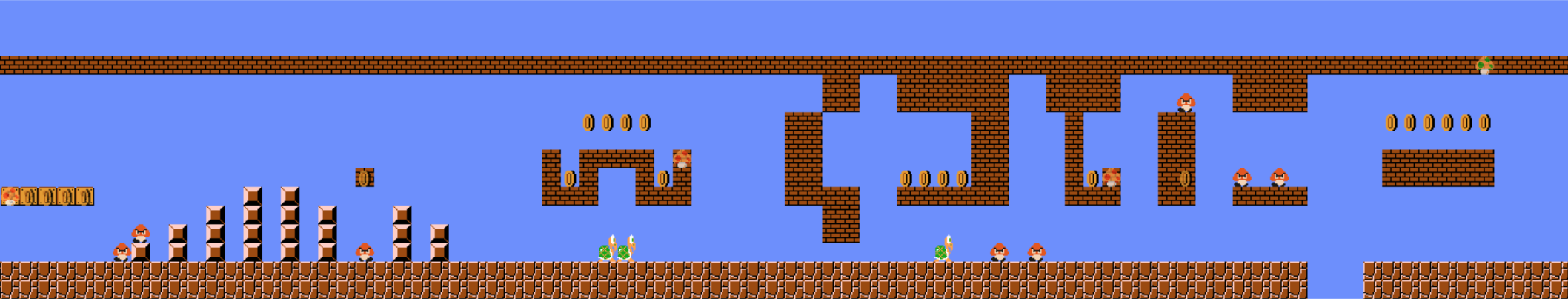}
    \caption{1-2 (original)}
\end{subfigure}
\hfill
\begin{subfigure}[b]{0.49\linewidth}
    \includegraphics[width=\linewidth]{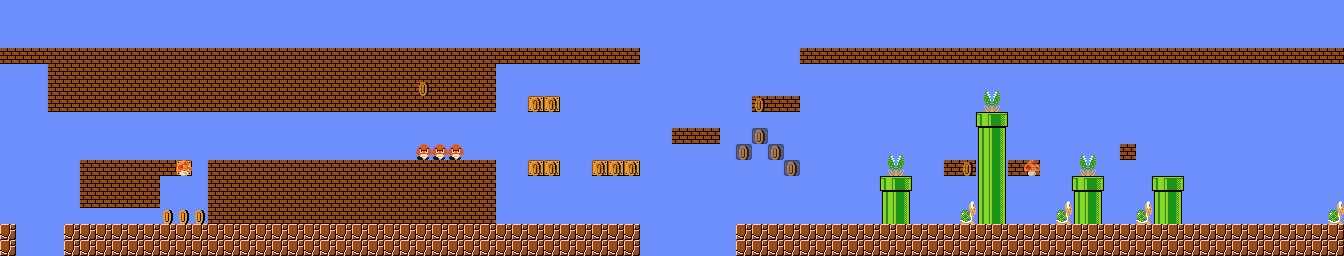}
    \caption{4-2 (original)}
\end{subfigure} \\
\begin{subfigure}[b]{0.49\linewidth}
    \includegraphics[width=\linewidth]{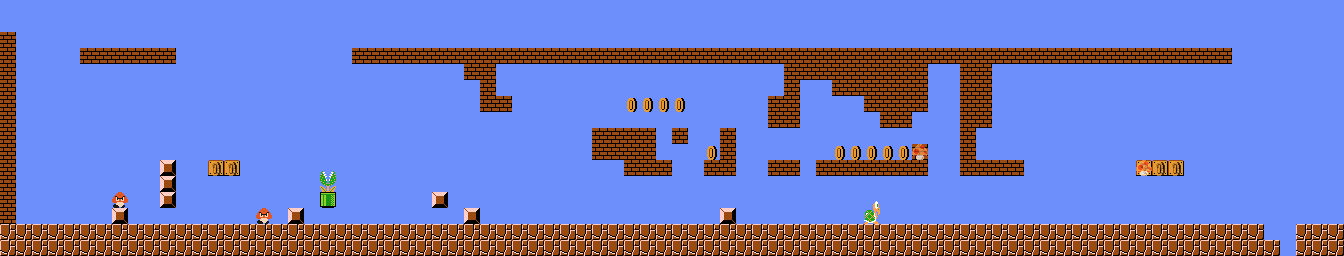}
    \caption{G 1-2 (ours)}
\end{subfigure}
\hfill
\begin{subfigure}[b]{0.49\linewidth}
    \includegraphics[width=\linewidth]{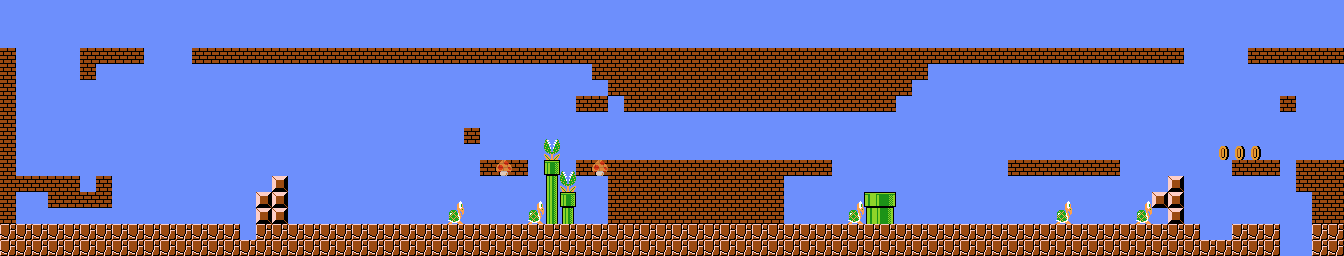}
    \caption{G 4-2 (ours)}
\end{subfigure}
\vspace{0.1cm}
\textbf{\small{Floating Platforms}}\\
\begin{subfigure}[b]{0.49\linewidth}
    \includegraphics[width=\linewidth]{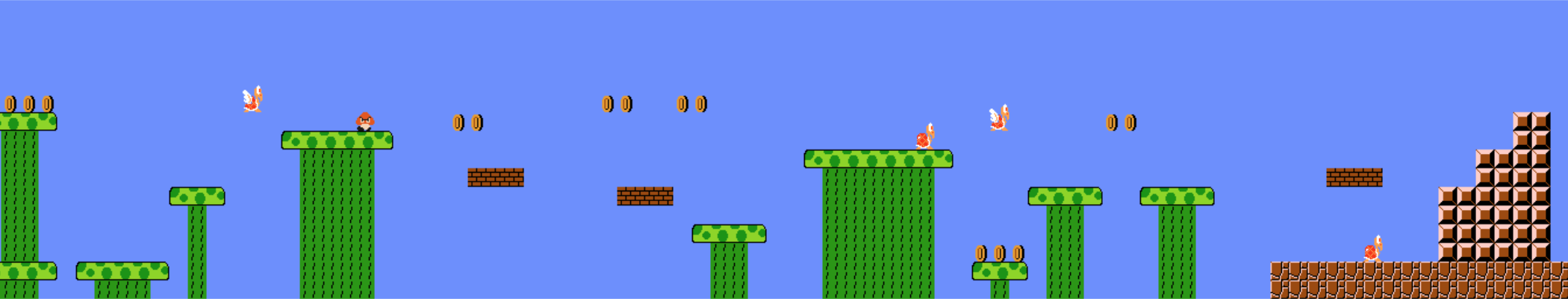}
    \caption{1-3 (original)}
\end{subfigure}
\hfill
\begin{subfigure}[b]{0.49\linewidth}
    \includegraphics[width=\linewidth]{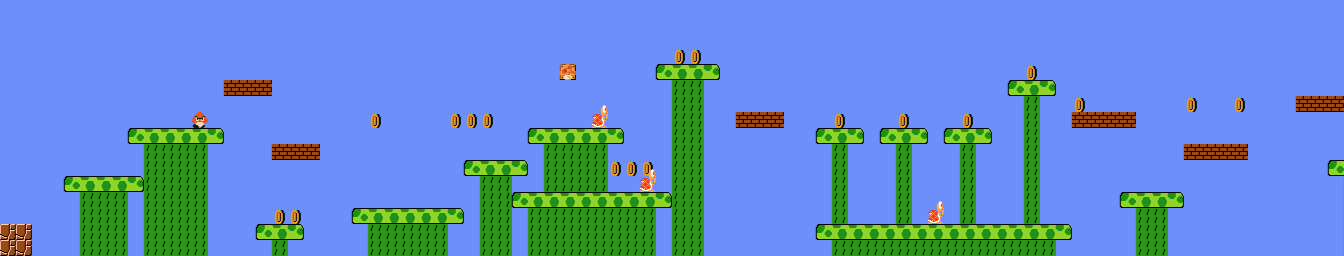}
    \caption{3-3 (original)}
\end{subfigure} \\
\begin{subfigure}[b]{0.49\linewidth}
    \includegraphics[width=\linewidth]{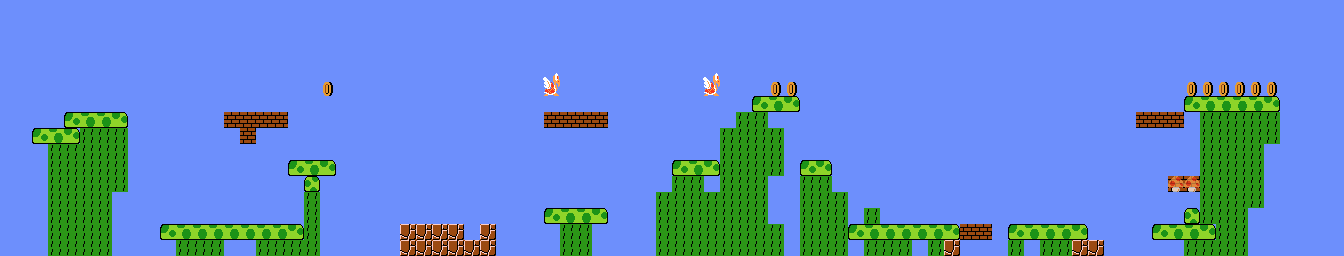}
    \caption{G 1-3 (ours)}
\end{subfigure}
\hfill
\begin{subfigure}[b]{0.49\linewidth}
    \includegraphics[width=\linewidth]{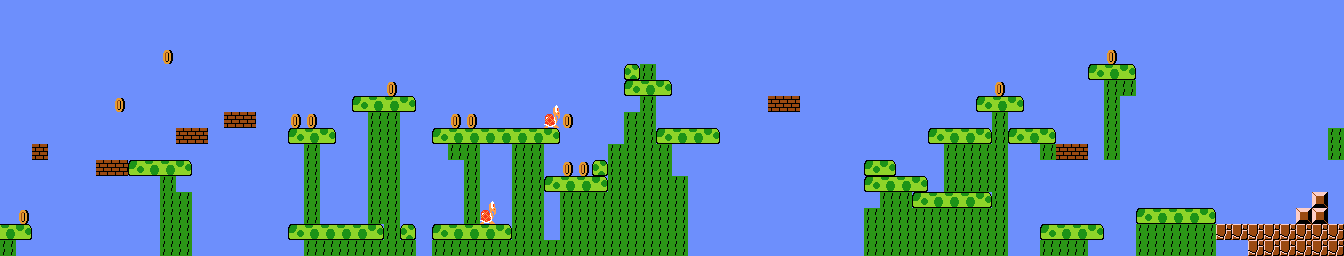}
    \caption{G 3-3 (ours)}
\end{subfigure}
\vspace{0.1cm}
\textbf{\small{Baselines}}\\
\begin{subfigure}[b]{0.49\linewidth}
    \includegraphics[width=\linewidth]{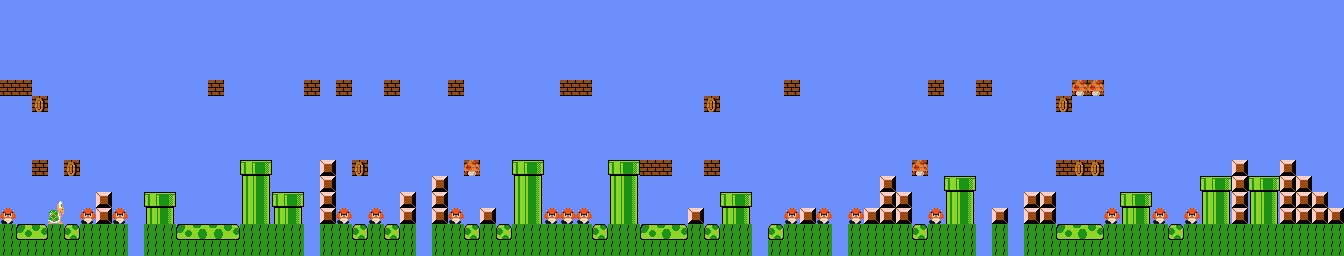}
    \caption{\citeauthor{dahlskogPatternsObjectivesLevel2013}}
\end{subfigure}
\hfill
\begin{subfigure}[b]{0.49\linewidth}
    \includegraphics[width=\linewidth]{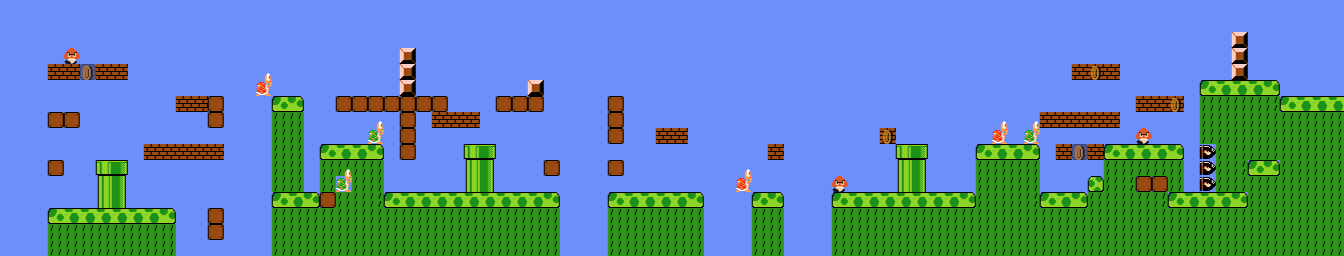}
    \caption{\citeauthor{shaker2010MarioAI2011}}
\end{subfigure} \\
\begin{subfigure}[b]{0.49\linewidth}
     \includegraphics[width=\linewidth]{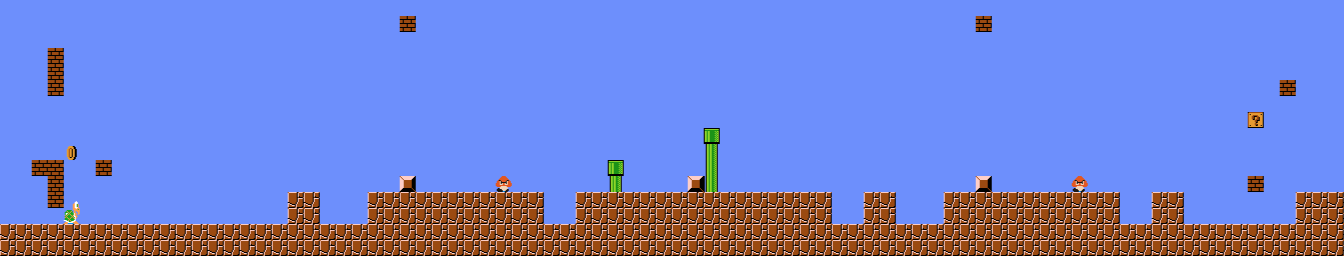}
    \caption{\citeauthor{greenMarioLevelGeneration2020}}
\end{subfigure}
\hfill
\begin{subfigure}[b]{0.49\linewidth}
    \includegraphics[width=\linewidth]{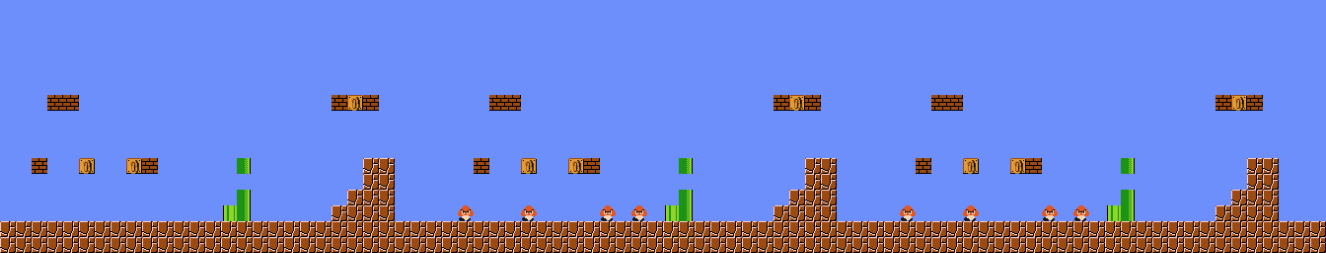}
    \caption{\citeauthor{volzEvolvingMarioLevels2018b}}
\end{subfigure}
\hfill
\end{center}
   \caption{
   Example levels generated by \methodname for the three different level types found in \acrlong*{SMB} in comparison to other generators. \methodname is able to capture the style of the level it was trained on.
   }
\label{fig:example_levels}
\end{figure}

Our experiments are split into two parts.
First, we perform a qualitative evaluation of the generated levels by presenting a number of samples to highlight capabilities of our approach.
Then, our generated levels are assessed with regards to their \gls*{TPKL-Div} \cite{lucasTilePatternKLDivergence2019} and visualised using an embedding of the level slices that is inspired by the \gls*{FID} \cite{heuselGANsTrainedTwo2018}.
In the second part, we show the generality of \methodname by applying it to \emph{Super Mario Kart} and present an example of level authoring.

We use the same hyperparameters for all experiments.
The samples were generated at scales $0.5, 0.75, 0.88$ and $1.0$ of the original training sample size.
The remaining hyperparameters and specific architectures will be published with our source code.

\subsection{Level Generation Evaluation}
\label{subsec:level_generation}

\subsubsection{Qualitative Examples}
\label{subsubsec:examples}

In this experiment, \methodname is trained on each of the levels provided by the \gls*{SMB} benchmark.
Fig.~\ref{fig:example_levels} shows randomly generated samples for levels of different types.
To increase comparability, all levels were cut (others), or generated (ours) to the same length.
The style of our generated samples matches that of the level they were trained on.
For example, the hidden 1-Up block in (b) is placed very similarly in (d).
However, the patterns in our samples are combined differently than in their training level (e.g. the three small platforms with coins from (j) are transformed to different heights in (l)).
All this while the general structure of the generated levels is similar to a \gls*{SMB} level.
We tested the validity of our generated content using the A* agent by Baumgarten \cite{togelius2009MarioAI2010}, who was able to win 65\% of randomly sampled levels compared to the 52\% of the original levels \cite{greenMarioLevelGeneration2020}.

While we focus on learning one generator for one level, other methods create one for all training levels.
The levels (m-p) are some example results of such generators. 
In them, different kinds of levels are mixed (e.g. pyramids are created on floating platforms) and the level style is not captured.
The levels (o) and (p) depict recognizable overworld levels, however the sample cut from (o) was trained on 4-2 and should therefore be more similar to an underground level.
The closest to a convincing overworld level is (p) which was also created using a GAN-based approach \cite{volzEvolvingMarioLevels2018b}.
However, this method relies on small samples that are stitched together and can result in repeating patterns. 

\subsubsection{Tile Pattern KL-Divergence}
\label{subsubsec:tile_pattern_kl_divergence}

\begin{figure}
\begin{center}
\includegraphics[width=\linewidth]{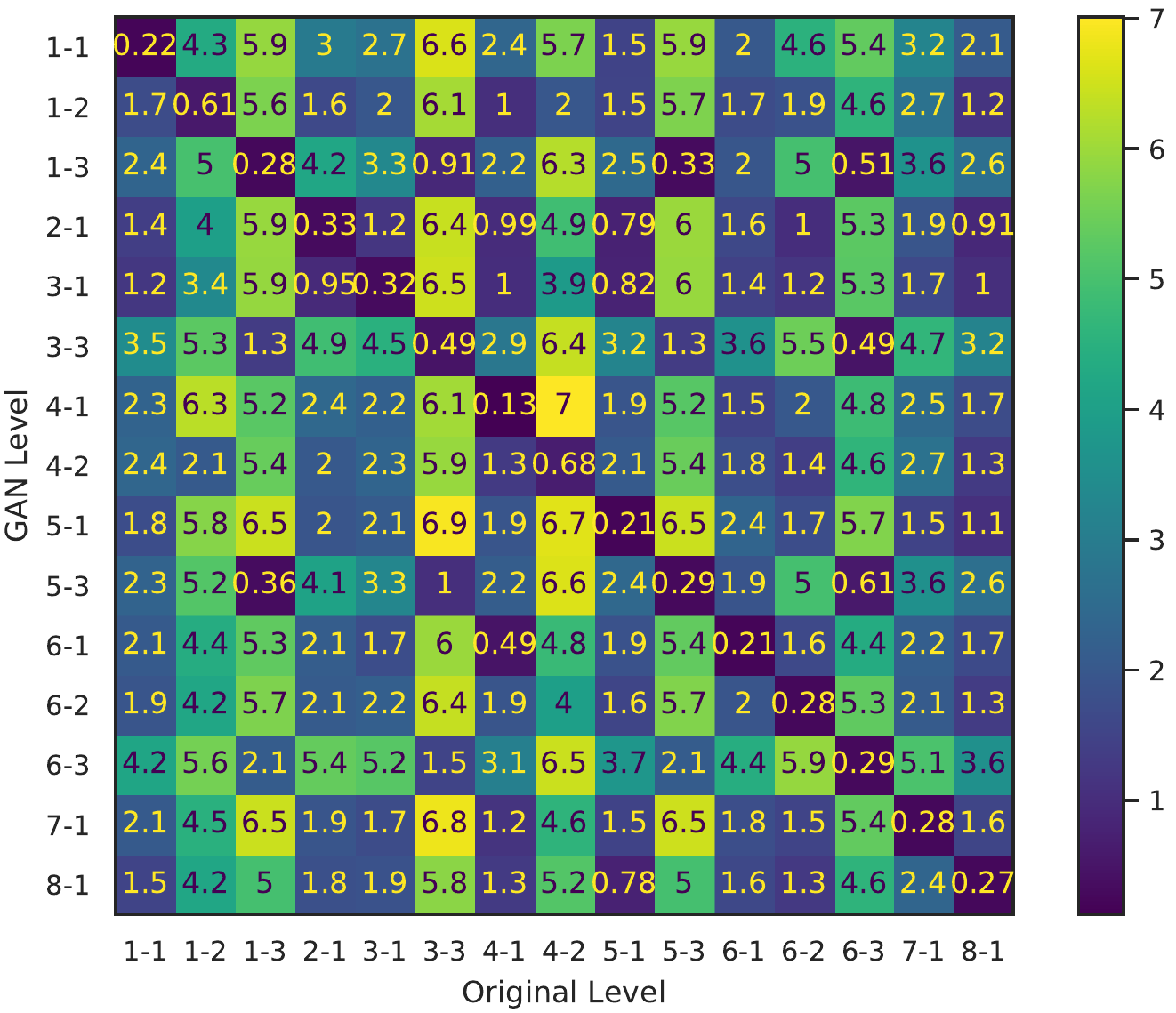}
\end{center}
   \caption{Mean Tile-Pattern-KL-Divergence between 100 generated levels and the original levels. Values are averaged over $2\times2$, $3\times3$ and $4\times4$ patterns. Each row represents a \methodname that was trained on the labelled \gls*{SMB} level.}
\label{fig:confusion_matrix}
\end{figure}

We use the \gls*{TPKL-Div} by \citet{lucasTilePatternKLDivergence2019} to evaluate the similarity of our generated patterns to the originals.
Fig.~\ref{fig:confusion_matrix} shows the results for all original \gls*{SMB} levels.
As expected, the values on the main diagonal, where the generated samples are compared to the level they were trained with, are very small.
This indicates that \methodname is able to model the original pattern distributions for any type of level. 
Also noticeable are spots in the matrix where a very low value occurs for a different level than the one trained on.
This happens because these levels are of a similar style.
For example, levels 1-3, 3-3, 5-3, and 6-3 are all levels with floating platforms, as shown in Fig.~\ref{fig:example_levels}(i)-(l).

\begin{table}
    \centering
    \caption{Average Tile Pattern KL-Divergence}
    \begin{tabular}{l  c}
    \toprule
        Algorithm & TPKL-Div. \\
    \midrule
        {ELSGAN \footnotemark[2]} & 1.58\\
        {GAN \footnotemark[2]} & 1.70\\
        {ETPKLDiv 3x3 \footnotemark[2]} & 0.88\\
        {Evolution World \footnotemark[3]} & 1.70\\
        {\methodname (ours)} & \textbf{0.33} \\
    \end{tabular}
\label{tab:kl-div-comparison}
\end{table}

Tab.~\ref{tab:kl-div-comparison} shows our resulting divergences compared to those reported by \citet{lucasTilePatternKLDivergence2019} and \citet{greenMarioLevelGeneration2020}. 
For our results, we computed the mean $2\times 2$, $3\times 3$, and $4\times 4$ pattern divergences with $w=1.0$.
As we train 15 separate generators, we average their values to get the result in Tab.~\ref{tab:kl-div-comparison}.
We generated 1000 sample levels with a size of $200 \times 16$ tokens for each generator.
On average, levels generated by \methodname produce a lower and therefore better \gls*{TPKL-Div}.
However, as the \gls*{TPKL-Div} measures only the differences for patterns already present in the original level, newly generated patterns are not taken into account.
Visual inspection of the generated levels (compare Figs.~\ref{fig:teaser} and \ref{fig:example_levels}) indicates that existing patterns are not only reproduced, but combined in novel ways and new patterns are generated.

\footnotetext[2]{Results by \citet{lucasTilePatternKLDivergence2019}, on level 1-1 averaged over $2\times 2$, $3\times 3$, and $4\times 4$ patterns with $w=1.0$}
\footnotetext[3]{Results by \citet{greenMarioLevelGeneration2020}, averaged over level 1-1, 4-2 and 6-1, only $3\times 3$ patterns}

Fig.~\ref{fig:mode_collapse} indicates that \glspl*{GAN} tend to produce very similar or even the same output.
We tested the variability of our generated content by computing the uniqueness of structures in our generated samples.
For that, we randomly picked 100000 square $16 \times 16$ slices evenly from our previously generated samples and found that an average of 90.62\% of them were unique.

\begin{figure*}
\begin{center}
\begin{multicols}{2}
\begin{subfigure}[b]{\linewidth}
    \includegraphics[width=.95\linewidth]{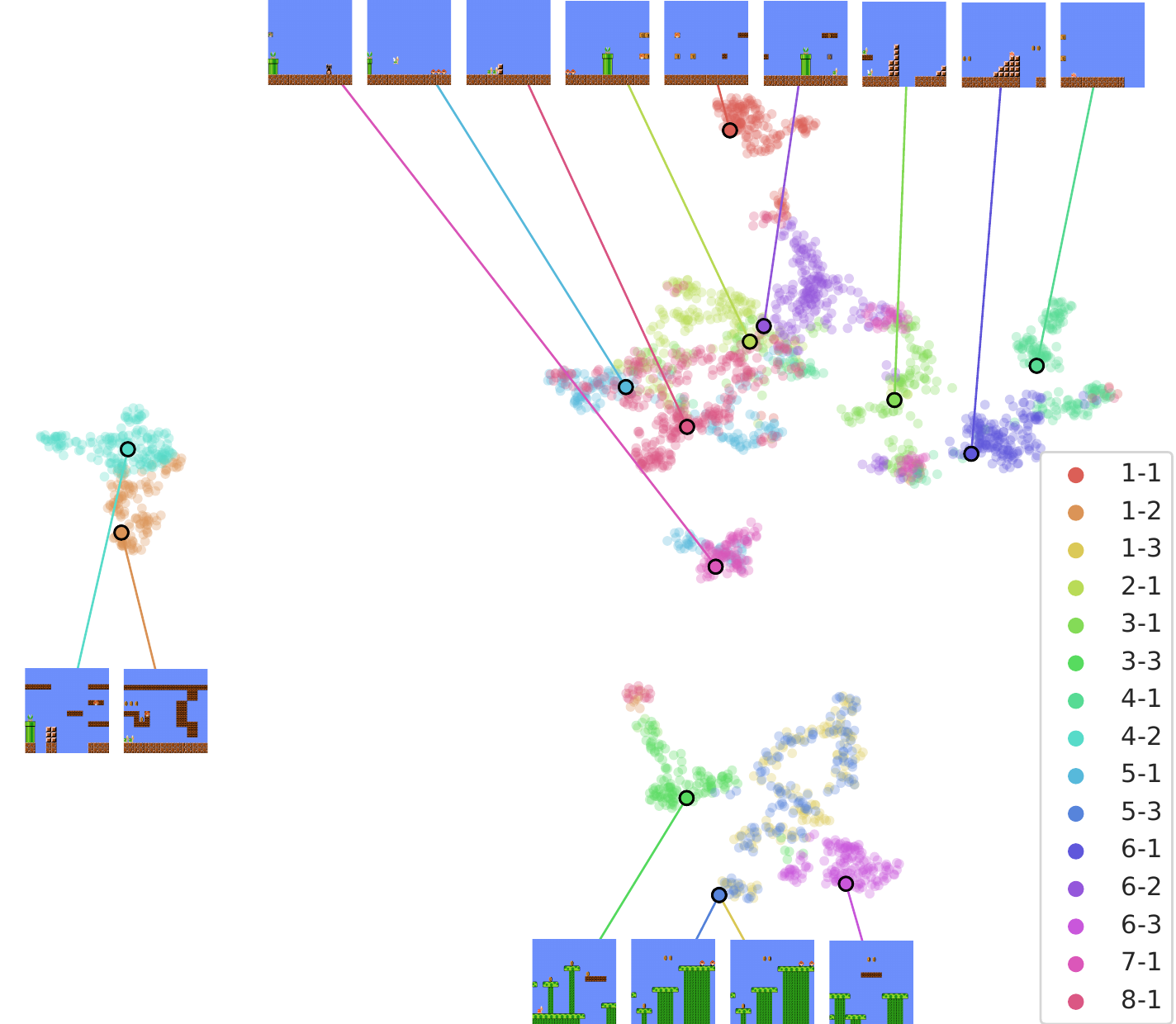}
    \caption{Original level slices}
\end{subfigure}
\begin{subfigure}[b]{\linewidth}
    \includegraphics[width=.95\linewidth]{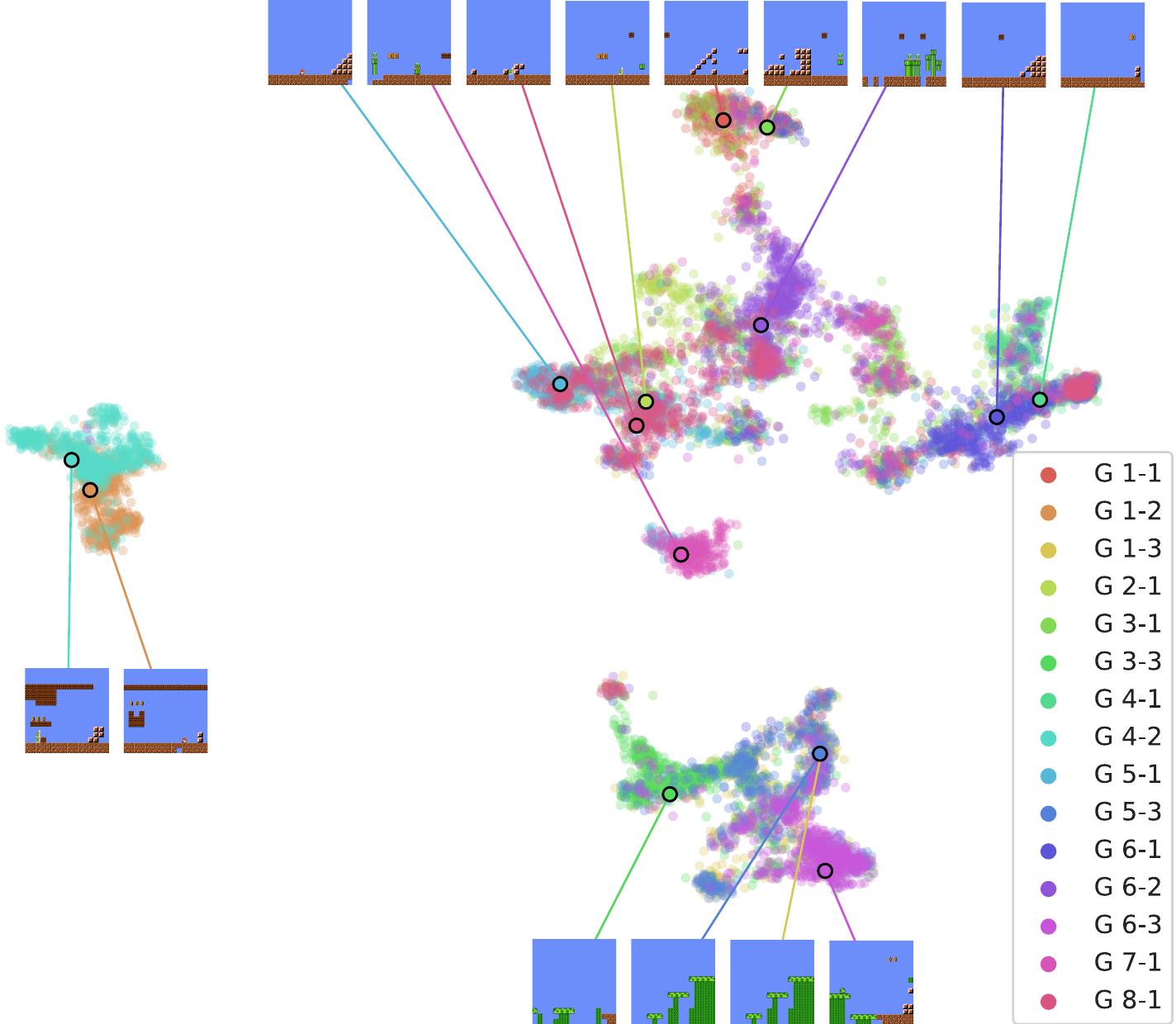}
    \caption{Our generated level slices (same transformation as (a))}
\end{subfigure}
\end{multicols}
\end{center}
   \caption{
   Level slice representations of a level classifier projected to two dimensions.
   Each point represents a $16 \times 16$ slice of a \acrlong*{SMB} level.
   The marked points are the ones closest to the mean of their respective level.
   For visualization purposes, a small amount of noise was added to the points, as some would otherwise overlap.
   The generated slices are close to the original slices of their respective level, with some slices being similar to other levels of the same style.
   }
\label{fig:slice_embeddings}
\end{figure*}

\subsubsection{Level Embeddings}
\label{subsubsec:level-embeddings}

Even though the \gls*{TPKL-Div} captures some aspects of the similarities between \gls*{SMB} levels, it is limited to patterns of fixed sizes.
We propose a new method that additionally results in an easily interpretable visualization.
Unlike the \gls*{TPKL-Div}, our distance metric is independent of the size of the patterns.
Similar to \gls*{FID}, we train a convolutional classifier $c$ on slices $s$ of the original levels to predict the level they are from.
\begin{equation}
    \text{level} \in \arg \max c(s) = \arg \max W\phi(s) + b
\end{equation}
The slice representation in the penultimate layer $\phi(s)$ of the classifier is a vector of non-linear features that is mapped linearly to the predicted level.

In Fig.~\ref{fig:slice_embeddings}(a), we visualize the distribution of these representations by projecting them to two dimensions with the \gls*{UMAP} \cite{mcinnesUMAPUniformManifold2018a} method.
This reveals the three level types which \gls*{SMB} levels fall into: Overworld to the top right, underground to the left and floating platforms at the bottom.

Fig.~\ref{fig:slice_embeddings}(b) shows the distribution of level slices generated by our 15 generators by mapping them with the same transformation learned in (a). 
Our generators are indeed generating slices very close but not limited to the manifold of their original level.
In some cases, our generators even learn to create slices that are not in their training distribution, but still within the level manifold.
Examples are G 1-3, G 5-3, and G 6-3 that generate slices in the same space as 3-3.
This experiment highlights the variability of the produced slices of a generator while still being within the intended constraints of being a \gls*{SMB} level of the same style.

\subsection{Level Authoring}
\label{subsec:level_style_transfer}

\begin{figure}
\begin{center}
\includegraphics[width=\linewidth]{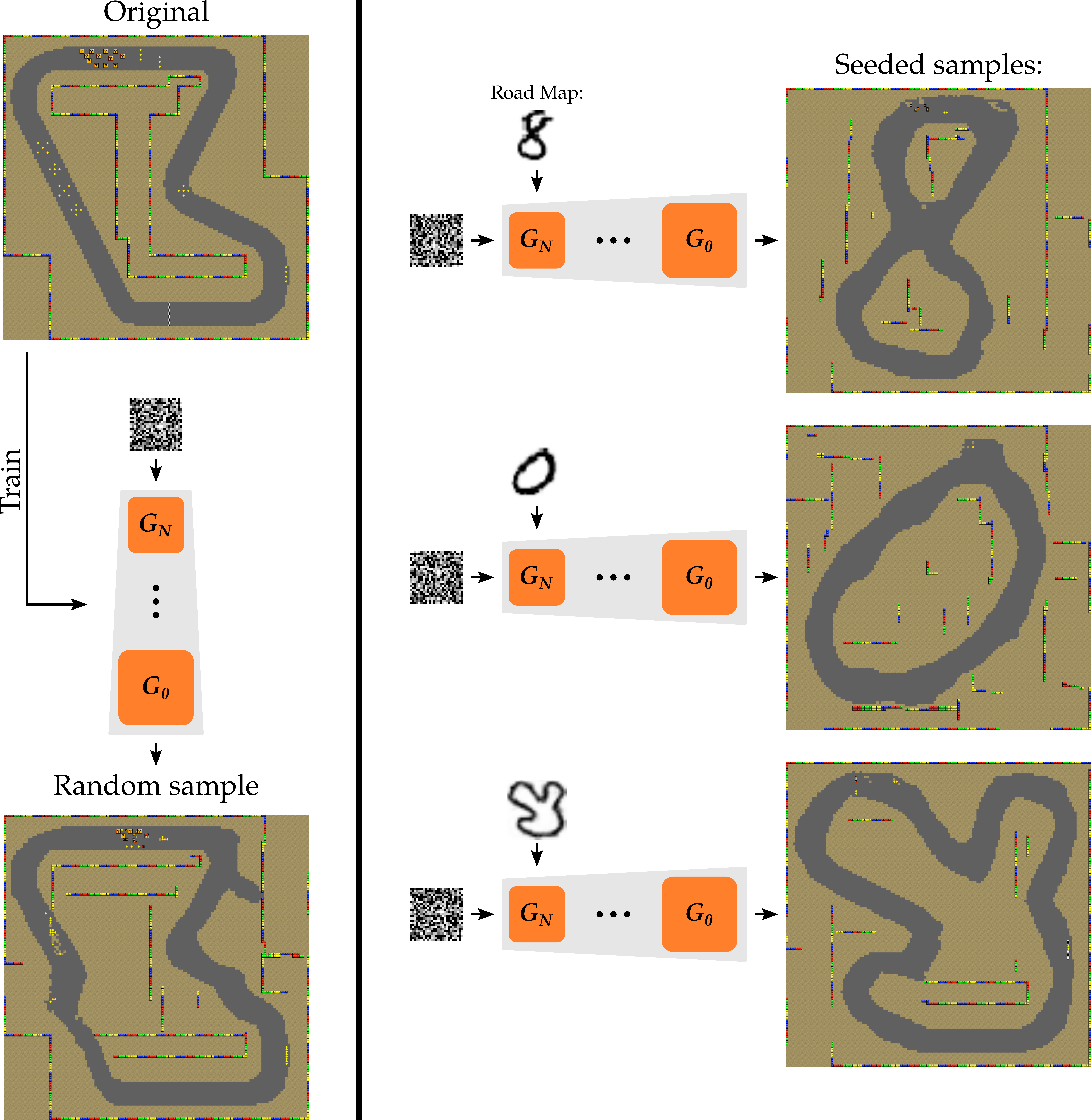}
\end{center}
   \caption{\methodname can be applied to levels of arbitrary token-based games, e.g. Super Mario Kart. 
   We can enforce a predetermined track layout by conditioning the generation process in the lowest scale.
   The examples shown are two different hand-drawn digits from the MNIST dataset \cite{lecun1998gradient} and an additional layout made for this example.
   }
\label{fig:seeding_examples}
\end{figure}

\methodname enables us to perform authoring of the global level structure.
This is made possible by editing the token maps in any of the scales, which results in the edit being represented in the generated level.
Fig.~\ref{fig:seeding_examples} shows examples of this application for seeding a track layout in Super Mario Kart.
\methodname is only trained on the original sample track and will generate track layouts similar to that. 
Because our method does not yet take playability into account, dead ends and unconnected track pieces can be generated.
Seeding a layout can not only ensure a connected and working racing track, it also allows the track to have a significantly different structure than the original sample.
Each seed can, depending on the noise in the other token maps and the other scales, generate an infinite amount of levels with the given structure. 
As the Super Mario Kart levels are much larger than the \gls*{SMB} levels, we used 5 convolutional layers instead of 3 and chose 9 scales ($0.2, 0.3, \dots, 1.0$). The token hierarchy is (from low to high) ground, wall, road and special (coins etc.).

\section{Conclusion and Future Work}
\label{sec:conclusion_and_future_work}

With this paper, we propose \methodname, a \acrlong*{PCG} algorithm that is able to generate token-based levels (as shown for \acrlong*{SMB} and Super Mario Kart) while being trained on a single example.
We expand on the novel SinGAN architecture to generate token-based levels instead of natural images.
The generated levels are evaluated qualitatively by computing their \acrlong*{TPKL-Div}.
Their visualization as slice embeddings offers a new way of comparing them with the original levels without specifying the pattern dimensions.
By seeding a predefined basic level layout, it is possible to generate new levels while still keeping the style that \methodname was trained on.
An example of this is shown by using hand drawn tracks for generating Super Mario Kart levels.
Our experiments demonstrate how \methodname is able to capture the patterns of its training input and generate consistent variations of it.

We intend to improve our approach in the future by also taking gameplay mechanics into account during the generation process.
Samples generated with \methodname are visually convincing \acrlong*{SMB} levels, but a proper study with human participants will help to assess the output quality in more depth.
Another future direction will be the application of \methodname to voxel-based games (e.g. Minecraft) or to maze games with a non-linear level structure.

\methodname is a step towards using \acrlong*{PCGML} during the game design process due to its low requirements for the amount of data necessary and its extension to Level Authoring.

\section{Acknowledgment}

This work has been supported by the Federal Ministry for Economic Affairs and Energy under the Wipano programme "NaturalAI" 03THW05K06.

\bibliography{bibliography}
\bibliographystyle{aaai}
\end{document}